\documentclass[letterpaper, 10 pt, conference]{ieeeconf}  

\IEEEoverridecommandlockouts                              

\usepackage[normalem]{ulem}
\usepackage[utf8]{inputenc}
\usepackage[T1]{fontenc}
\usepackage[textsize=scriptsize, disable]{todonotes}
\usepackage{lipsum}
\usepackage{algorithm}
\usepackage{algpseudocode}
\usepackage{url}
\usepackage{multirow}
\usepackage{blindtext}
\usepackage[separate-uncertainty = true, free-standing-units, space-before-unit, use-xspace]{siunitx}
\usepackage{color,soul}
\usepackage{svg}
\usepackage{amsmath} 
\usepackage{amsfonts}
\usepackage{amssymb}  
\usepackage[nolist]{acronym}
\usepackage{listings}
\usepackage{float}

\begin{acronym}
\acro{api}[API]{application programming interface}
\acro{bt}[BT]{behavior tree}
\acroplural{bt}[BTs]{behavior trees}
\acro{dt}[DT]{digital twin}
\acroplural{dt}[DTs]{digital twins}
\acro{ebt}[eBT]{extended behavior tree}
\acroplural{ebt}[eBTs]{extended behavior trees}
\acro{fsm}[FSM]{finite state machine}
\acroplural{fsm}[FSMs]{finite state machines}
\acro{gui}[GUI]{graphical user interface}
\acroplural{gui}[GUIs]{graphical user interfaces}
\acro{htn}[HTN]{hierarchical task network}
\acro{mes}[MES]{manufacturing execution system}
\acroplural{mes}[MES]{manufacturing execution systems}
\acro{mg}[MG]{motion generator}
\acroplural{mg}[MGs]{motion generators}
\acro{btmg}[BTMG]{behavior tree and motion generator}
\acroplural{btmg}[BTMGs]{behavior trees and motion generators}
\acro{owl}[OWL]{Web Ontology Language}
\acro{pddl}[PDDL]{Planning Domain Definition Language}
\acro{rdf}[RDF]{Resource Description Framework}
\acro{rl}[RL]{reinforcement learning}
\acro{ros}[ROS]{Robot Operating System}
\acro{wm}[WM]{world model}
\acro{ee}[EE]{end effector}
\end{acronym}

\newfloat{lstfloat}{tbp}{lop}
\floatname{lstfloat}{Listing}

\definecolor{lightgray}{rgb}{0.98,0.98,0.98}
\definecolor{commentgreen}{RGB}{2,112,10}
\definecolor{eminence}{RGB}{108,48,130}
\definecolor{weborange}{RGB}{255,165,0}
\definecolor{frenchplum}{RGB}{129,20,83}
\definecolor{ceruleanblue}{rgb}{0.16, 0.32, 0.75}

\newcommand{\insertref}[1]{\todo[color=green!40]{#1}}

\title{\LARGE \bf
Using Knowledge Representation and Task Planning for Robot-agnostic Skills on the Example of Contact-Rich Wiping Tasks
}

\author{Matthias Mayr$^{1}$, Faseeh Ahmad$^{1}$, Alexander Duerr$^{1}$ and Volker Krueger$^{1}$
\thanks{$^{1}$Department of Computer Science, Faculty of Engineering (LTH), Lund University, SE~221~00 Lund, Sweden. E-mail: <firstname>.<lastname>@cs.lth.se.}
}

\begin{document}
\maketitle
\thispagestyle{empty}
\pagestyle{empty}

\begin{abstract}
The transition to agile manufacturing, Industry 4.0, and high-mix-low-volume tasks require robot programming solutions that are flexible. However, most deployed robot solutions are still statically programmed and use stiff position control, which limit their usefulness.

In this paper, we show how a single robot skill that utilizes knowledge representation, task planning, and automatic selection of skill implementations based on the input parameters can be executed in different contexts. We demonstrate how the skill-based control platform enables this with contact-rich wiping tasks on different robot systems. To achieve that in this case study, our approach needs to address different kinematics, gripper types, vendors, and fundamentally different control interfaces. We conducted the experiments with a mobile platform that has a \textit{Universal Robots UR5e} 6 degree-of-freedom robot arm with position control and a 7 degree-of-freedom \textit{KUKA iiwa} with torque control.

\end{abstract}

\section{Introduction}
The need for flexible and skill-based robot control systems in industrial robot applications is becoming increasingly important. With the rise of automation and the growing complexity of tasks, robots must be able to adapt to changing environments and perform a variety of tasks without lengthy configuration times. This requires a control system that can respond quickly and accurately to changes in the environment and the task at hand. However, most of the deployed systems are statically programmed and perform repetitive tasks.
This introduces the need for a flexible and skill-based robot control system in industrial robot applications that provides flexibility and supports agile manufacturing.

\begin{figure}[tpb]
	{
		\setlength{\fboxrule}{0pt}
		\framebox{\parbox{3in}{
				\centering
				\includegraphics[width=0.98\columnwidth]{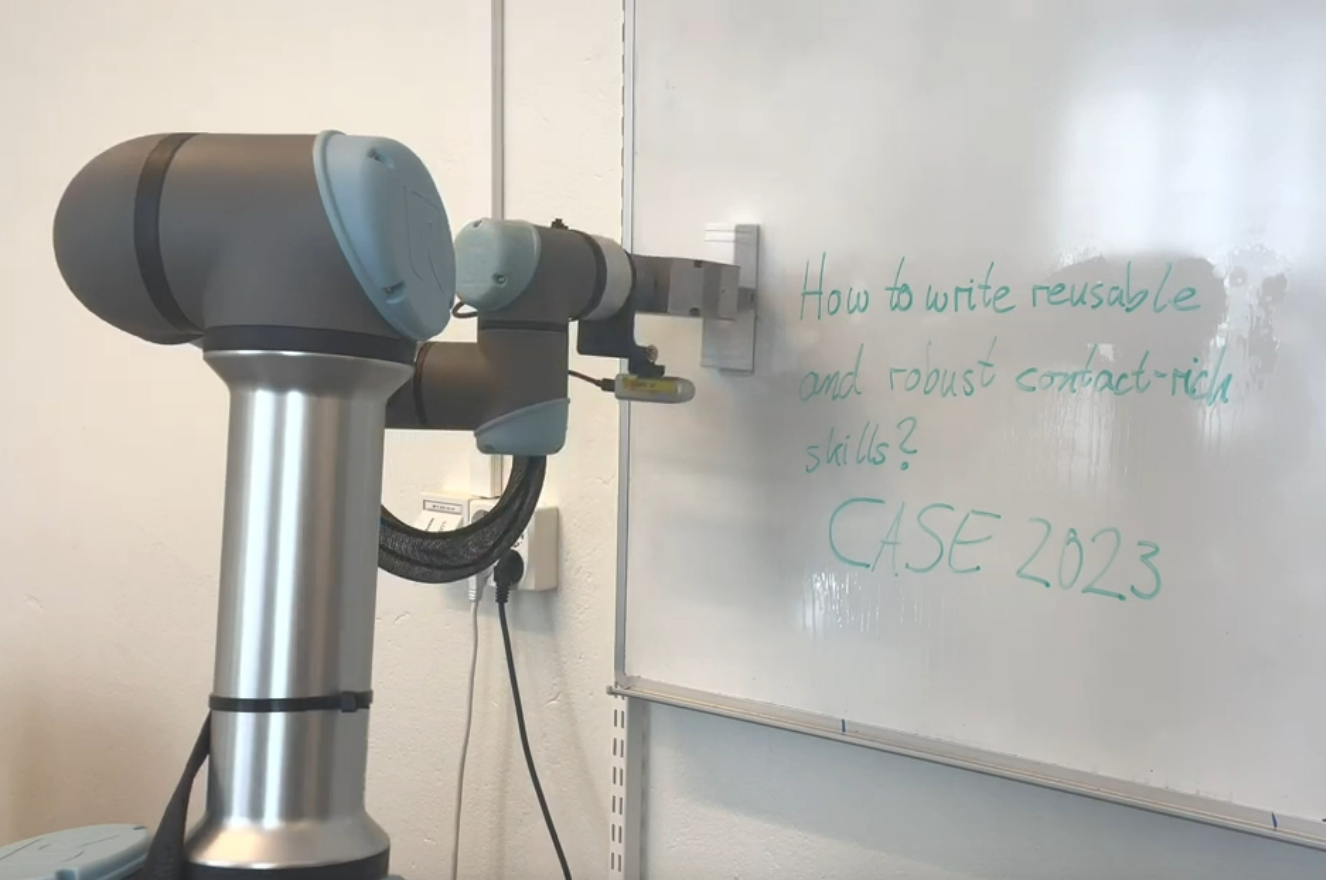}
            }
		}
	}
	\caption{With the proposed abstractions and knowledge representation, several different robot systems can execute the same skills while still using their respective hardware such as grippers and controllers.}
	\label{fig:overview}
\end{figure}

First platforms for skill-based systems such as \textit{ClaraTy}~\cite{volpe2001claraty} or \textit{LAAS}~\cite{bensalem2009designing}, paved the way. In the area of knowledge integration frameworks, the system built around the \textit{Rosetta ontology} \cite{stenmark13faiaa, stenmark15racm} and \textit{Knowrob}~\cite{tenorth092iicirs, tenorth13tijorr} created a solid foundation. The latter addresses the service domain, while the former is placed in the context of industrial robotics.
To transfer skills from one robot to another is nontrivial if robots of different types and vendors are involved.
For instance,~\cite{topp2018ontology} demonstrated skill transfer for contact-free motions within a single family of robots that share the same vendor-supplied high-level interface \textit{ABB RAPID}.
In that paper, the simulated experiments were conducted with high-gain position control. This is, of course, still a standard practice within many tasks that have little variation and high certainty.
However, the increasing number of contact-rich tasks that manipulators should be able to solve~\cite{wuthier2021productive} indicates a clear need for flexible and compliant systems.
Achieving compliant behaviors heavily depends on the hardware and vendor, and there are no standard interfaces available for this purpose.
In this paper, we explore the question of how skills can be transferred between collaborative robots, such as the \emph{Universal Robot} robots and the \emph{KUKA iiwa} or the \emph{Franka Emika Robot (Panda)}. To do this, we develop a suitable representation of the available robot hardware, a knowledge infrastructure to maintain this hardware knowledge, and a suitable compliant controller that works across different types of robot arms. To validate and verify our claims, we test our architecture on a whiteboard wiping task, as shown in Fig. \ref{fig:overview}, using a \emph{UR5e} and a \emph{KUKA iiwa}. The whiteboard wiping task is an example of a typical contact-rich task that requires a compliant controller. For this, the \emph{UR5e} has a force-torque sensor on the wrist, while robots such as the \emph{iiwa} and the \emph{Panda} have torque sensors in their joints.

In summary, our paper makes the following contributions:  
\begin{enumerate}
    \item We present a complete architecture that represents, plans, and executes skills such that they can be transfered between robots.
    \item We demonstrate these ideas based on a whiteboard wiping task using a \emph{UR5e} and a \emph{KUKA iiwa} robot.
    \item All essential building blocks, including the impedance controller implementations are available as open-source software.
    \item The code is written in a robot-agnostic way, making the results and insights scalable to diverse collaborative robotic hardware and tasks.
\end{enumerate}

\section{Related Work}
\subsection{Robot Skill Systems}
We can classify skill-based architectures based on their target domain. Projects like \textit{CAST}~\cite{wyatt2010self} and \textit{KnowRob}~\cite{tenorth092iicirs, tenorth13tijorr} are for service robotics and general platforms are \textit{ClaraTy}~\cite{volpe2001claraty}, \textit{LAAS}~\cite{bensalem2009designing} and \textit{SmartMDSD}~\cite{stampfer2016smartmdsd}.
However, platforms that address the specific needs of industrial robotics are the most relevant for this work. \textit{CoStar}~\cite{paxton2017costar} has shown to integrate multiple robot systems and provides a \ac{gui} for the design and the monitoring of skills. However, it lacks explicit knowledge representation and the possibility to perform task planning.
The use of ontologies to describe skills in robotic applications has been explored in various areas, including medical and surgical robotics~\cite{bruno2019caresses, gibaud2018toward}, perception ontologies~\cite{azevedo2018ontpercept} for robots, and more~\cite{manzoor2021ontology}. Among these,~\cite{stenmark13faiaa, stenmark15racm, topp2018ontology} present a knowledge-based approach for programming synchronized motions between robotic systems and human-robot interactions that is particularly relevant for our work.

In~\cite{topp2018ontology}, the authors demonstrate the effectiveness of their approach through experiments that involve transformations of dual arm programs and the transfer of skills between robot systems with different kinematics. Their approach uses finite state machines to implement skill behavior and generates \textit{ABB RAPID} code that is specific to \textit{ABB} robot controllers. Our work differs in that we use behavior trees for skill behavior implementation and our implementation is not limited to a particular robot controller.~
In addition,~\cite{topp2018ontology} was validated only in simulation, and the tasks were not contact-rich.

\subsection{Robot Motions}
\label{robot-motions}
In the literature, Movement Primitives (MPs) have been a reliable tool for generating robot motions. Dynamic Movement Primitives (DMPs)~\cite{976259,ijspeert2002learning} and Probabilistic Movement Primitives (ProMPs)~\cite{paraschos2013probabilistic} have been successfully used to generate arm motions. Both DMPs and a motion representation based on behavior trees and motion generators (BTMG) that will be introduced in Section~\ref{sec:btmg} use attractor landscapes to reach a target location. DMPs are non-linear dynamical systems that allow trajectory control, and, like our BTMGs, they have a non-linear forcing term that enables the modification of the original trajectory. However, in contrast to DMPs, the parameters of BTMGs can be specified explicitly, giving us a better understanding of the robot motion and allowing for a better intregration with knowledge representation. 

\textbf{Wiping Applications:} In~\cite{leidner2015classifying, leidner2016robotic, leidner2019cognition} the authors discuss research on robotic manipulation of wiping tasks using artificial intelligence reasoning methods. ~\cite{leidner2015classifying} proposes a classification of compliant manipulation tasks based on symbolic effects, subcategorizes wiping tasks, and demonstrates how to concretize actions based on the example of shards sweeping with a broom. \cite{leidner2016robotic} investigates the reasoning and action execution problems involved in the execution of wiping tasks and proposes a high-level abstraction representation of wiping tasks to develop generalized action execution mechanisms. \cite{leidner2019cognition} combines reasoning methods and compliant robotic manipulation to solve wiping tasks by introducing a qualitative particle distribution model, an approach to generate whole-body wiping motions based on effect-oriented policies, an approach to assess wiping motion quality, and the semantic interpretation of contact situations.
Rather than achieving the best wiping skill, the focus of our paper is on presenting a complete architecture that represents, plans, and executes skills, and then using wiping as a challenging example to demonstrate these ideas.

\section{Transferable Robot Skills}
\label{sec:skills}
In this section we introduce the prerequisites that enable to transfer skills between different robot systems. In addition to the skill platform \textit{SkiROS2} and its skill model, the necessary aspects of knowledge presentation and task planning are introduced. The control implementations and motion generation are introduced in Section~\ref{sec:control}.
\subsection{SkiROS2}
\label{sec:skiros2}
\begin{figure}[tpb]
	{
		\setlength{\fboxrule}{0pt}
		\framebox{\parbox{3in}{
        \centering
		\includegraphics[width=0.9\columnwidth]{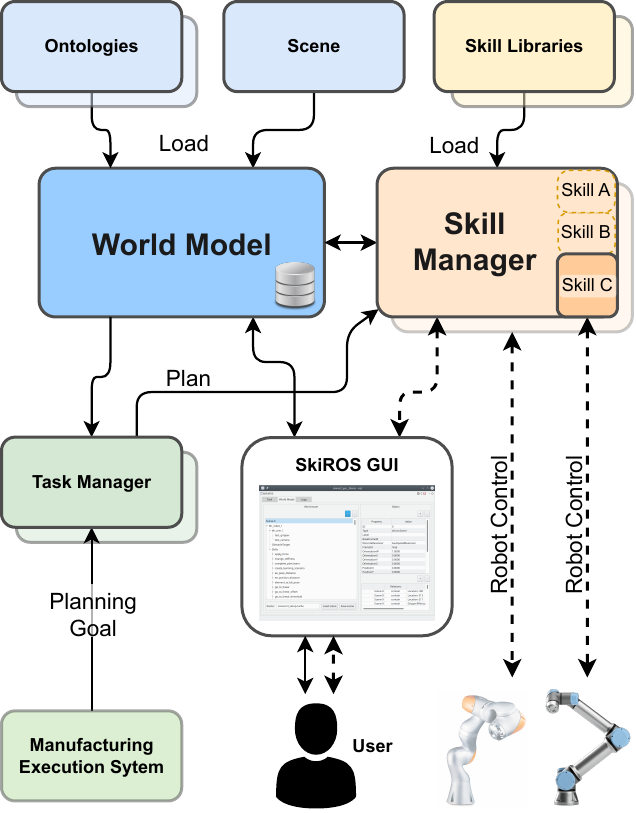}}
        }
	}
	\caption{An outline of the SkiROS2 architecture. The world model stores the knowledge about the relations, environment and the skills. The skill manager loads and executes the skills. Dashed lines show control flows and solid lines show information flows. Shaded blocks indicate possible multiple instances.}
	\label{fig:architecture}
\end{figure}
We utilize the skill-based robot control platform \textit{SkiROS2}~\cite{rovida2017extended} to implement transferable robot skills. It is used in several research projects for motion generation~\cite{rovida2018motion}, robot coordination~\cite{wuthier2021productive} and task-level planning with \ac{rl}~\cite{mayr2022combining, mayr2022skill, mayr22priors, ahmad2023learning}.
The architecture is shown in Fig.~\ref{fig:architecture} and the main components are the skill manager and the world model. The \ac{wm} contains the knowledge about the world and represents it in a \ac{rdf} database. Ontologies such as the \textit{Suggested Upper Merged Ontology (SUMO)} describe the available concepts such as objects or skills and their relations to each other. In addition to that, a \textit{scene} contains the concrete instances of the concepts that are defined in the ontologies. Such instances can be the present robot systems, workstations or objects that are currently known.

The skill manager is responsible for loading a set of specified skills from the skill libraries. Each skill that is successfully loaded gets a semantic description in the \ac{wm}. This includes the parameters and pre-, hold- and post-conditions. In addition to that, the skill manager provides services to start, pause and stop skills and to manage their execution.

The task manager accepts the planning goals in \ac{pddl}~\cite{aeronautiques1998pddl, rovida2017extended}. Such goals can come from a user or an external entity such as a \ac{mes}. An example of such a goal is to place an object at a specific location: \texttt{(skiros:contain skiros:Location-1 skiros:Product-1)}. The task manager automatically creates a \ac{pddl} planning domain from the knowledge in the \ac{wm}. The currently available skills are added with their preconditions and effects, and the relevant instances in the scene are automatically added to a planning problem. Then a \ac{pddl} planner is called to find a sequence of skills that is guaranteed to be optimal~\cite{eyerich2009using}. Finally, the plan is converted back into a \ac{bt}, automatically expanded with the relevant skill implementations and grounded with the instances in the \ac{wm}~\cite{rovida2017extended}.

\subsection{Skill Model}
\label{sec:skill-model}
\begin{figure}[tpb]
	{
		\setlength{\fboxrule}{0pt}
		\framebox{\parbox{3in}{
				\centering
			\includegraphics[width=0.9\columnwidth]{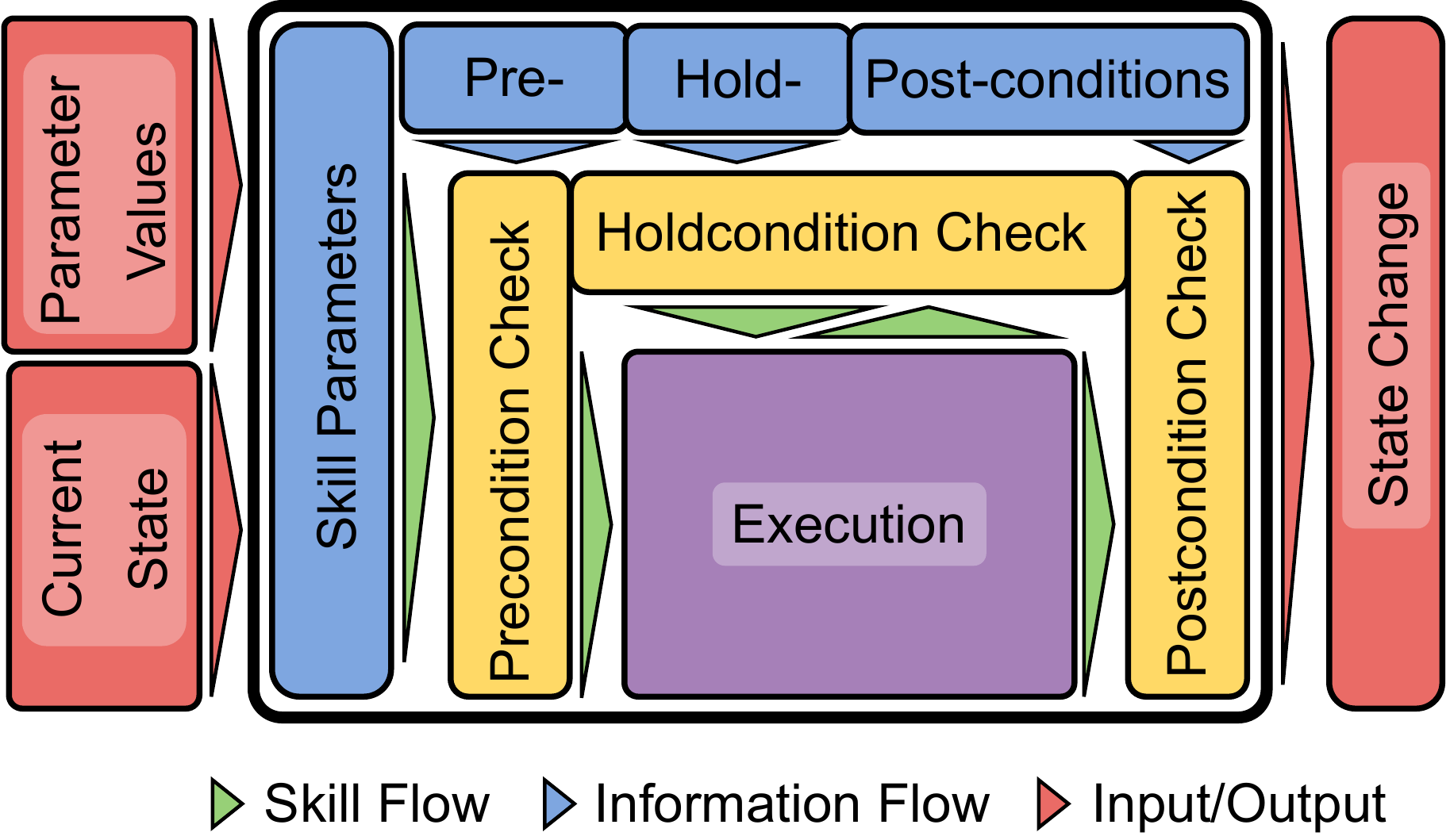}
		}}
	}
	\caption{The conceptual model of a skill in SkiROS2. \mbox{Pre-,} and hold-conditions ensure that the skill is only executed in the correct world state. Post-conditions check if the desired changes have been achieved.}
	\label{fig:skill}
\end{figure}

We define a skill as a parametric procedure that changes the world from some initial state to some new state~\cite{bogh2012does}. It is shown in Fig.~\ref{fig:skill}. A skill consists of two main components: a skill description and the skill implementation. A skill description contains the skill parameters and the necessary pre-, hold- and post-conditions. These define what prerequisites need to exist to execute a skill and state the expected effects of running it. In the task manager, these conditions can be used for automatic task planning. Furthermore, the parameters state the input and output of a skill. There are three different types of parameters: 1) required, 2) optional and 3) inferred. An example of a required parameter of a "pick" skill is to specify the robot arm, while an optional parameter could be to slow down the execution for testing. The inferred parameters are reasoned about when starting the skill. An example is a "gripper" parameter for a pick skill that already has the arm as a required parameter. With a precondition we can define a relation between both of them and we can use the knowledge in the \ac{wm} to specify the inferred parameter automatically~\cite{rovida2017extended}.

In addition to the skill description that describes the semantic actions of a skill, the skill implementation is a concrete version for the execution. Each skill implementation implements exactly one description with its parameters and pre-, hold- and post-conditions. There are two complexities of skills: Primitive skills that are semantically atomic and compound skills. Primitive skills provide a Python interface for the initialization, start, execution and stop of a skill. In contrast to that, compound skills allow to connect an arbitrary number of primitive skills and compound skills in a \ac{bt}~\cite{colledanchise17bt, rovida2017extended}.
A skill implementation is also allowed to change the skill description, such as adding addition conditions or modifying existing conditions. This is important to support multiple, specific skill implementations. Examples of this are gripper actuation skills. A basic skill description can have a parameter "gripper" of the \ac{wm} type \texttt{Element("rparts:GripperEffector")} and a Boolean parameter "opening state". The implementation of such a skill depends on the actual gripper model and a specific implementation can set the parameter to a specific subtype of \texttt{rparts:GripperEffector}, such as \texttt{scalable:RobotiqGripper}.

\subsection{Knowledge Representation and Task Planning}
An important aspect is the separation of skills and tasks. Skills are written in a parametric way to be used in various applications and different tasks. This separation is achieved by explicitly storing knowledge in the \ac{wm}. Since the conditions and task-specific knowledge is not implicitly represented in the skills, this allows for a modular and extensible design that easily allows to add scenarios or robot systems at a later point.
The \ac{wm} stores this knowledge in a semantic \ac{rdf} database that utilizes the \ac{owl}. SkiROS2 introduces its own ontology, which contains the necessary concepts and relations for skill modeling and reasoning. It is based on the \textit{Core Ontology for Robotics and Automation (CORA)} that is in the \textit{IEEE Standard Ontologies for Robotics and Automation (IEEE Std 1872™-2015)}\insertref{ieee standard}.

An example for a part of the semantic description of a robot arm is in Listing~\ref{lst:robot}. The listing contains the relevant knowledge for the trajectory generation and motion execution. This includes, but is not limited to properties such as \ac{ros} interfaces of a robot and the relevant aspects of a motion planning configuration. In addition there are elements needed for reasoning about relations such as the attached gripper.

\begin{lstfloat}
    \begin{lstlisting}[language=Python, basicstyle=\scriptsize, label={lst:robot}, caption={An excerp from the semantic description of one of the robots used in the case study. It includes the necessary knowledge to parameterize motion skills and perform spatial reasoning.\vspace{0.2cm}}]
scalable:Ur5-2 a scalable:Ur5, owl:NamedIndividual ;
    rdfs:label "scalable:ur5"^^xsd:string ;
    skiros:BaseFrameId "cora:Robot-1"^^xsd:string ;
    skiros:CartesianGoalAction "/cartesian_trajectory_generator/goal_action"^^xsd:string ;
    skiros:OverlayMotionService "/cartesian_trajectory_generator/overlay_motion"^^xsd:string ;
    skiros:CartesianStiffnessTopic "/cartesian_param_filter/stiffness_goal"^^xsd:string ;
    skiros:CartesianWrenchTopic "/cartesian_param_filter/force_goal"^^xsd:string ;
    skiros:CompliantController "cartesian_compliance_controller"^^xsd:string ;
    skiros:JointConfigurationController "scaled_pos_traj_controller"^^xsd:string ;
    skiros:DiscreteReasoner "AauSpatialReasoner"^^xsd:string ;
    skiros:FrameId "scalable:Ur5-2"^^xsd:string ;
    skiros:LinkedToFrameId "ur5e_base_link"^^xsd:string ;
    skiros:MotionExe "/scaled_pos_traj_controller/follow_joint_trajectory"^^xsd:string ;
    skiros:MoveItGroup "manipulator"^^xsd:string ;
    skiros:MoveItReferenceFrame "ur5e_base_link"^^xsd:string ;
    skiros:MoveItTCPLink "ur5e_tcp_link"^^xsd:string ;
    skiros:hasA scalable:WsgGripper-3 .
    \end{lstlisting}
\end{lstfloat}

Tasks are executed by the skill manager and there are two ways to start them: 1) Select and parameterize a skill manually. For example in the \ac{gui} or from an \ac{api} or 2) by providing a planning goal to the task manager and executing the skill sequence. Tasks are concrete skill (sequence) instances with a specific goal. Within a task, the skills can utilize a blackboard to exchange information with their parameters.
\section{Control and Motion Generation}
\label{sec:control}

\subsection{Behavior Trees and Motion Generators}
\label{sec:btmg}
Behavior Trees (BT) are mathematical models used for plan representation and execution. They have been shown to perform reliably in games~\cite{colledanchise142iicirsa}, artificial intelligence and robotics~\cite{iovino2020survey}.
A BT is an acyclic graph defined with nodes and edges defined in a typical parent-child relationship. The nodes are of two types: 1) \textit{control flow} and 2) \textit{execution}. The \textit{control flow} nodes are responsible for directing the flow of the tree and are traditionally classified on the basis of their operation as: 1) sequence, 2) selector, 3) parallel and 4) decorator ~\cite{colledanchise17bt, marzinotto142iicrai}. In essence, a \ac{bt} works by means of sending a tick signal from the \textit{Root} node. The signal then moves through the nodes controlled by the \textit{control flow} nodes. The return statements from the \textit{execution} nodes are \emph{running},  \textit{success} or \textit{failure}. In order to combine BTs with task-level planning,~\cite{rovida2017extended} proposes \textit{extended BTs}.

A motion generator (MG)~\cite{rovida2018motion} generates an arm motion using impedance controllers to control the end effector (EE) in Cartesian space. A MG follows a simple attractor landscape that uses a virtual spring to attract the EE to the desired location, whereas virtual dampers allow safe motion by slowing down the overall motion. In addition, it also allows for a deviation by superimposing a motion on top of the original trajectory. For example, in~\cite{mayr2022skill} we used a Cartesian linear trajectory superimposed by an Archimedes spiral to solve the peg-in-hole task.

A Behavior Tree and Movement Generator (BTMG) combines the strengths of both BT and MG to model robotic skills. A BTMG is a parameterized representation that allows us to specify not only the structure of BT but also the properties of the MG. In essence, the BTMG representation not only covers the aspects of plan representation and execution, but also determines the arm motions for the execution strategies~\cite{ahmad2022generalizing}.

\subsection{Trajectory Generation}
\label{sec:motion}
The concept of the MG has been introduced in~\cite{rovida2018motion}, which consists of both a trajectory generation method and a corresponding controller to execute the generated trajectory on the manipulator. In the following section, we provide a description of the trajectory generation process adopted for this project.

In many industrial settings, operations are performed in a Cartesian task space rather than in the joint space of the robot. Therefore, we argue that motion references are best generated and provided in the Cartesian end-effector space. In this work, we adopt Cartesian linear paths\footnote{https://github.com/matthias-mayr/cartesian\_trajectory\_generator} to generate the corresponding trajectories. This is achieved by selecting appropriate acceleration profiles and setting the maximum Cartesian translational and rotational velocities. Furthermore, we offer the option to synchronize motions that have both translational and rotational components. Specifically, the trajectory generator takes the new goal pose as input and outputs end-effector pose references, which are then sent to the corresponding controller.

Aside from specifying a new goal pose, it is also possible to overlay additional motions on top of the reference pose. This capability has been utilized in various studies such as~\cite{mayr21iros, mayr2022combining, mayr2022skill}, where an Archimedes spiral is used to find a hole in an insertion procedure. In this work, we apply a sine motion to improve the wiping performance of the robot.
\subsection{Compliant Control}
\label{sec:compliant-control}
Manipulation tasks that have uncertainties need compliant control solutions to provide the necessary flexibility. Such uncertainties can come from either the task itself or inaccurate knowledge about the placement of objects. Another source could be the placement of the robot, which becomes relevant when working with a mobile robot. The robot systems used in the case study provide different interfaces for control commands and therefore also have different solutions for compliant control.

\subsubsection*{Cartesian Impedance Control}
\label{sec:impedance-control}
Robot systems such as the \textit{KUKA iiwa} or the \textit{Franka Emika Robot} offer an interface to send commanded torques for each of the joints. This can be used to perform Cartesian impedance control~\cite{mayr2022c++} which allows compliant control of the end effector in task space.

Torque-controlled robots are typically gravity compensated. The rigid-body dynamics of such a system in the joint space is described as $q\in  \mathbb{R}^{n}$ \insertref{springer 2016}:
\begin{equation}
    M(q)\ddot{q} + C(q,\dot{q})\dot{q} = \tau_{\mathrm{c}} + \tau^{\mathrm{ext}}
    \label{eq:rigid-body}
\end{equation}
where $M(q)\in  \mathbb{R}^{n\times n}$ is the generalized inertia matrix, $C(q,\dot{q})\in  \mathbb{R}^{n\times n}$ captures the effects of Coriolis and centripetal forces, $\tau_{\mathrm{c}}\in  \mathbb{R}^{n}$ represents the input torques, and $\tau^{\mathrm{ext}}\in  \mathbb{R}^{n}$ represents the external torques, with $n$ being the number of joints of the robot.

Moreover, the torque signal commanded by this controller to the robot, $\tau_{\mathrm{c}}$ in Equation (\ref{eq:rigid-body}), is composed of the superposition of three joint-torque signals:
\begin{equation}\label{eq:tau_c}
    \tau_{\mathrm{c}} = \tau_{\mathrm{c}}^\mathrm{ca} + \tau_{\mathrm{c}}^\mathrm{ns} + \tau_{\mathrm{c}}^\mathrm{ext}
\end{equation}
where

$\tau_{\mathrm{c}}^\mathrm{ca}$ is the torque commanded to achieve a Cartesian impedance behavior with respect to a Cartesian pose reference in the $m$-dimensional task space, $\xi^{\mathrm{D}}\in\mathbb{R}^{m}$, in the frame of the end effector of the robot:
\begin{equation}\label{eq:tau_sup}
    \tau_{\mathrm{c}}^\mathrm{ca} = J^{\mathrm{T}}(q)\left[-K^\mathrm{ca}\Delta \xi-D^\mathrm{ca}J(q) \dot{q}\right]
    \end{equation}
with $J(q)\in \mathbb{R}^{m \times n}$ being the Jacobian relative to the end-effector (task) frame of the robot, and $K^\mathrm{ca}\in \mathbb{R}^{m \times m}$ and $D^\mathrm{ca}\in \mathbb{R}^{m \times m}$ being the virtual Cartesian stiffness and damping matrices, respectively.

$\tau_{\mathrm{c}}^\mathrm{ns}$ is the torque commanded to achieve a joint impedance behavior with respect to a desired configuration and projected in the null-space of the robot's Jacobian, to not affect the Cartesian motion of the robot's end effector.

$\tau_{\mathrm{c}}^\mathrm{ext}$ is the torque commanded to achieve the desired external force command in the frame of the end effector of the robot, $F_{\mathrm{c}}^\mathrm{ext}$:
    \begin{equation}\label{eq:tau_ext}
        \tau_{\mathrm{c}}^\mathrm{ext} = J^{\mathrm{T}}(q)F_{\mathrm{c}}^\mathrm{ext}
    \end{equation}

In this work we utilize $\tau_{\mathrm{c}}^\mathrm{ext}$ and $\tau_{\mathrm{c}}^\mathrm{ca}$ in combination with changing stiffnesses by passing configurations through skill primitives.

\subsubsection*{Cartesian Compliance Control}
In contrast to the \textit{KUKA iiwa}, the \textit{Universal Robots UR5e} like many other industrial manipulators does not support direct torque control. This means that a compliant control solution is needed that can work with joint position or joint velocity commands. Recently, such a method was introduced through forward dynamics compliance control~\cite{scherzinger2017}. It combines admittance, impedance and force control into one control strategy. The control loop is closed only through a force-torque sensor, which makes it system independent. Internally it utilizes a forward dynamics simulation of a virtual model of the robot system to map Cartesian inputs to joint commands. For further details, we refer the interested reader to~\cite{scherzinger2017} and~\cite{scherzinger2019inverse}. Like the impedance controller that was introduced in the previous section, this controller solution also allows for the runtime modification of applied external forces, reference poses and stiffness changes. We utilize these capabilities when modeling behaviors with BTs.

\section{Case Study: Wiping Task}
We demonstrate the usage of skills in different tasks and on different robot system with the example of a wiping task.

\begin{figure}[tpb]
	{
		\setlength{\fboxrule}{0pt}
		\framebox{\parbox{3in}{
				\centering
                \vspace{0.1cm}
				\includegraphics[width=0.9\columnwidth]{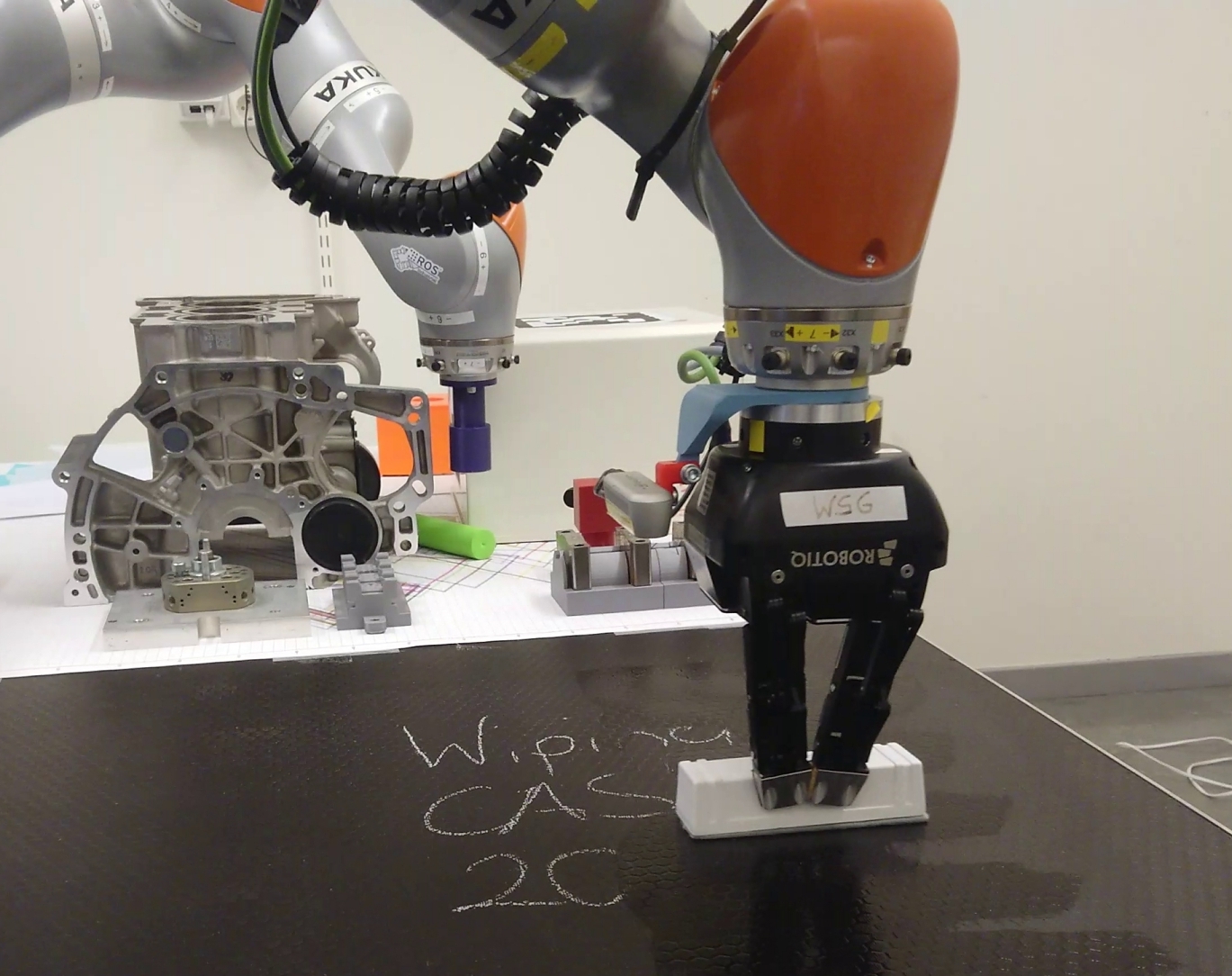}
            }
		}
	}
	\caption{The \textit{iiwa} wiping task on a table. The different hardware needs a different controller setup and gripper skills.}
	\label{fig:iiwa}
\end{figure}

\begin{figure*}[tpb]
	\centering
	{
		\setlength{\fboxrule}{0pt}
		\framebox{\parbox{0.98\textwidth}{
		\centering

		\includegraphics[width=\textwidth]{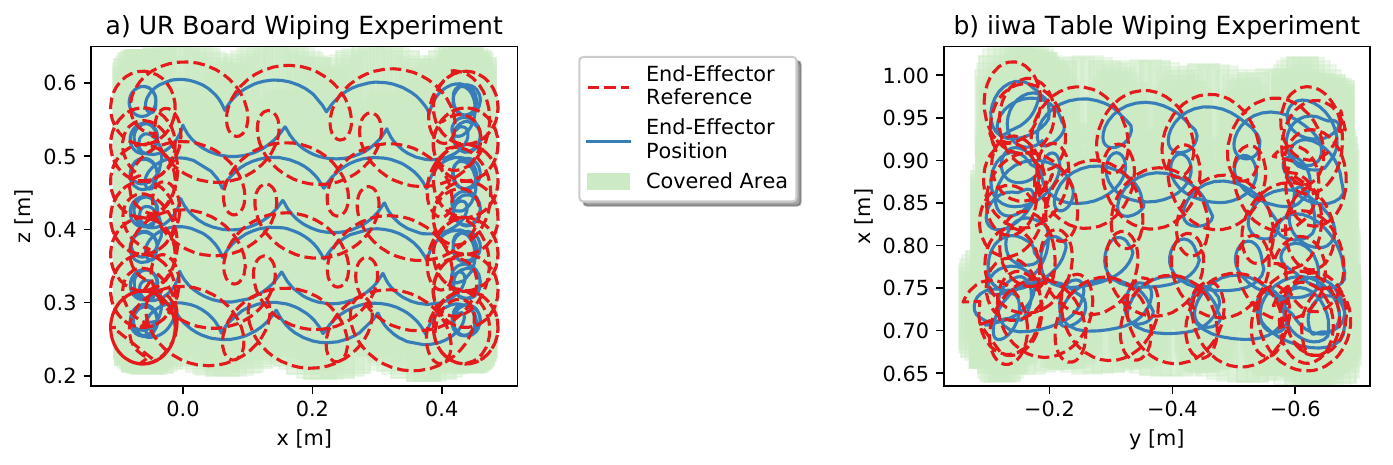}
	\vspace{-0.7cm}
	\caption{The resulting reference positions and actual positions of the end effectors in a) the whiteboard wiping task and with the \emph{UR5e} b) the table wiping task with the \emph{iiwa}. The shaded green shows the surface area that has been covered by the eraser that is used for cleaning.}
	\label{fig:exp}
	}}}
\end{figure*}

\subsection{Challenges}
The task at hand and the robot systems selected for this study present a range of intriguing challenges. Firstly, the task instances involve surfaces with different properties, namely a smooth whiteboard and a rough industrial table. The whiteboard is mounted to a wall, whereas the table surface is parallel to the floor. Moreover, we assume that the precise distance between the robot and the surface may not be known in advance. The robot systems are equipped with different grippers, each of which has its own communication interface. Finally, the robots have distinct kinematics, vendor-supplied programming solutions, and control interfaces. In this context, we demonstrate how a skill that utilizes knowledge representation, planning and the automatic skill implementation selection can be executed in different context on different robot systems.

\subsection{Implementation}
The implementation of such a skill incorporates several key aspects that are discussed in~\cite{rovida2018motion}. Specifically, we utilize a \ac{mg} that comprises of both trajectory generation~(Sec.\ref{sec:motion}) and a controller (Sec.\ref{sec:compliant-control}).

As inputs for the skill, only the robot arm and a surface to clean need to be selected among the available instances in the \ac{wm}. All other necessary data, such as robot-specific properties (e.g., the controller to use) and the surface properties (e.g. the dimensions, force to apply) are automatically fetched from the \ac{wm} as part of the knowledge integration. Apart from the parameters, this skill also contains the necessary pre- and postconditions for planning. The integrated plannerin SkiROS can receive a goal such as \texttt{(skiros:clean scalable:Workstation-1186 scalable:Cell-12)}. For instance, one precondition of the skill is that the tool must already be held in the gripper. This is leveraged in task-level planning, where the robot is required to pick up the tool first. In case of a mobile system, this may entail driving to a different location. The postcondition of the task is that the surface is cleaned.

To begin the wiping process, the end effector is moved to the corner of the surface to be cleaned. Once there, it disables the stiffness of the compliant controller along the normal of the surface and applies a predetermined force. In our experiments, \si{8}{\newton} are specified for surfaces that are easier to clean, such as the whiteboard, and \si{18}{\newton} for the table. The robot then applies an overlay motion, in this case a circular motion, while performing the wiping action.
From there the skill executes laps by moving the robot's end-effector to the right, then down, and finally left until the entire surface is covered. Finally, upon completion of the task, the robot stops the force application, increases the stiffness along the surface normal, and moves the end-effector away from the surface.

\subsection{Experiments and Discussion}
We conducted the experiments using two different robot systems.  The first system is a mobile platform that has a \textit{UR5e} 6 degree-of-freedom robot arm with a \textit{Schunk} two finger gripper. The second system is a 7 degree-of-freedom \textit{KUKA iiwa} with a \emph{Robotiq} 3-finger gripper. We selected two different surfaces for the cleaning task, each with different orientations and surface properties. 

The resulting reference path and the actual path for the \emph{UR5e} are shown in Fig.~\ref{fig:exp}a). The implementation spends more time in the left and right extremes since the sub-trajectory ends there and it switches to new lane.  This can be adressed by blending between trajectories. During the execution we also noticed subtle vibrations, especially when the arm was stretched out.

In our second experiment we used the \textit{KUKA iiwa} robot system to wipe the surface of a table. As in the previous experiment, the important attributes were represented in the \ac{wm}, and the skill could be started by simply selecting the surface and the robot arm. The reference path and the actual path for the robot end effector during the task is shown in Figure~\ref{fig:exp}b). The controller is very stable and there are no vibrations. We observed that the eraser was occasionally lifted slightly during the wiping process. This could potentially be addressed by increasing the rotational stiffness or applying more pressure to the surface.

In both tasks, the respective robot systems were able to completely clean the surface with the proposed implementation. In principle, the surface can also be cleaned using only the Cartesian linear motion. However, as intuitively expected, the addition of an additional overlay motion improves the cleaning performance in practice.

\section{Conclusions and Future Work}
The transition towards Industry 4.0 brings challenges such as small batch sizes, constantly changing tasks and environments.
Together with the transition towards human-robot colaboration and handling contact-rich tasks, it creates the need for platforms that can address these challenges.

In this paper we presented a complete architecture that allows to transfer skills between different robot systems. We explain the necessary prerequisites and knowledge representation. We demonstrated this by successfully performing contact-rich wiping tasks with a \emph{KUKA iiwa} and a \emph{Universal Robots UR5e}. All essential building blocks including the skill-based system and the controllers are available as open-source software and the results are expected to scale to other hardware and tasks as well.

While the wiping performance was sufficient in our tasks, it can potentially be increased by learning a good combination of path velocities and applied force. A visual inspection with a camera is a next inuitive step. We are also looking into an evaluation with other platforms such as the \textit{Franka Emika Robot (Panda)}.

\section*{ACKNOWLEDGMENT}
This work was partially supported by the Wallenberg AI, Autonomous Systems and Software Program (WASP) funded by Knut and Alice Wallenberg Foundation.


\bibliography{root}
\bibliographystyle{bib/IEEETran}

\end{document}